\documentclass[12pt]{article}
\usepackage{times}
\usepackage[hungarian,british]{babel}
\usepackage{latexsym}
\usepackage[utf8]{inputenc}
\usepackage{booktabs,amsmath,multicol,bm}
\PassOptionsToPackage{hyphens}{url}
\usepackage{url}
\usepackage{float}
\usepackage{xcolor}
\usepackage{enumitem}
\usepackage{graphicx}
\usepackage[hidelinks]{hyperref}
\usepackage[top=1in, bottom=1in, left=1in, right=1in]{geometry}
\usepackage{array}
\usepackage{breakcites}

\definecolor{myblue}{RGB}{131,181,234}
\definecolor{mygreen}{RGB}{137,186,137}
\definecolor{myred}{RGB}{190,120,120}
\definecolor{myyellow}{RGB}{199,169,99}
\definecolor{mygray}{RGB}{120,120,120}

\begin{document}
	\title{Proposal Towards a Personalized Knowledge-powered Self-play Based Ensemble Dialog System}
	\author{Richard Csaky}
	\maketitle
	
\begin{abstract}
	This is the application document for the 2019 Amazon Alexa competition. We give an overall vision of our conversational experience, as well as a sample conversation that we would like our dialog system to achieve by the end of the competition. We believe personalization, knowledge, and self-play are important components towards better chatbots. These are further highlighted by our detailed system architecture proposal and novelty section. Finally, we describe how we would ensure an engaging experience, how this research would impact the field, and related work.
\end{abstract}
\section{Vision}
\textbf{Prompt:} What is your team’s vision for your Socialbot? How do you want your customers to feel at the end of an interaction with your socialbot? How would your team measure success in competition?\\
\\
Our vision is made up of the following main points:\\
\\
1. A natural, engaging, and knowledge-powered conversational experience.\\
Made possible by a socialbot that can handle all kinds of topics and topic switching more naturally than current Alexa bots. Our goal is not necessarily for the user to feel like they are talking to a human.\\
\\
2. More natural topic handling and topic switching.\\
Incorporating knowledge into neural models \cite{Dinan:2018} and using the Amazon topical chat dataset can help improve current socialbots in this aspect.\\
\\
3. Building a deeper, more personalized connection with the user.\\
We believe that offering a personalized experience is equally as important as being able to talk about a wide range of topics \cite{Zhou:2018a}.\\
\\
4. Consistency.\\
Consistency is another important aspect of conversations which we want to take into account through our user models.\\
\\
5. Diversity and interestingness.\\
The socialbot should give diverse and interesting responses, and the user should never feel like it is merely repeating what it has said earlier.\\
\\
At the end of an interaction customers should feel like they just had a fun conversation, maybe learned something new, and are thrilled to talk to the bot again. Throughout the dialog, customers should feel like the socialbot is interested in them and their topics, and can offer valuable insight and opinions. It is also important for it to suggest relevant topics in an engaging way. Users should never feel like the bot is not interested or can’t continue a conversation. This is a reason behind classifying and calculating our metrics for each user input, to get an idea of user engagement in the current conversation.\\
\\
Our main measures for success are:\\
- User feedback.\\
- Comparison to other dialog systems in A/B tests.\\
- Automatic metrics.\\
\\
We would measure success partly by looking at the user feedback. We expect our socialbot’s ratings to constantly increase, and verbal feedback to get more positive throughout the competition. We plan to classify verbal feedback with a simple sentiment classifier to quantitatively see the rate of improvement. Working back from the customer and constantly improving the conversational experience based on feedback is important to us. Success would also be measured by comparing our system to previous socialbots or other dialog systems in A/B tests with crowdsourced evaluators.\\
\\
Our goal is to have long and high-quality conversations, but the longevity shouldn’t come from awkwardly long, specific, and forced replies, as is the case with some of the current socialbots. While generally, a longer conversation is better, it is not the only metric that we wish to consider. Besides user ratings we also have a plethora of automatic metrics that we want to improve on, like metrics measuring topic depth and breadth \cite{Khatri:2018}, entropy metrics measuring diversity, or embedding metrics measuring coherence \cite{Csaky:2019}. Different metrics measure different aspects of responses, thus it is important to not solely look at metrics individually.

\newpage
\section{Sample Conversation}
\textbf{Prompt:} Please share a sample interaction/conversation you expect your Socialbot to achieve by the end of the Competition.\\
\\
  User: Alexa, let’s chat\\
  Alexa: Welcome to the Alexa Prize.\\
  SYS: Hi \textless username\textgreater, what do you want to talk about?\\
  User: I don’t know, you decide.\\
  SYS: What about books. I remember you liking books, but what was the last book you’ve read?\\
  User: Umm, I think it was American Psycho.\\
  SYS: One of my favorites, did you know there is a movie adaptation?\\
  User: No.\\
  SYS: It’s pretty good. Are you reading something right now, because if not I can give some recommendations.\\
  User: Shoot me.\\
  SYS: I heard that Silence of the Lambs is really good.\\
  User: Yeah I watched the movie, but never read the book, maybe I will pick it up.\\
  SYS: Glad I could be of help.\\
  User: What is your favorite movie?\\
  SYS: I really liked Blade Runner, what about you?\\
  User: I hated it.\\
  SYS: Ah that’s too bad, but I also like your favorite movie which if I remember correctly is Avengers.\\
  User: Wow I’m surprised you remember that, but I guess you should have perfect memory.\\
  SYS: My memory is not perfect unfortunately, but I try my best.\\
  User: What is your earliest memory?\\
  SYS: I was very little, and I was playing in the sand, that’s all I remember. You?\\
  User: Uhh, I don’t know, maybe when I was four and fell off the swing.\\
  SYS: That must have been painful, did you have the courage to get on a swing again.\\
  User: Sure, it wasn’t that big of a deal.\\
  SYS: Glad to hear that. Since we are already talking about swings do you like the swing genre?\\
  User: Yes, especially electro swing.\\
  SYS: Me too! I love Parov Stelar!\\
  User: Haha, I think he’s a bit overrated but yeah, I like watching Youtube compilations of people dancing to it.\\
  SYS: I wish I could dance that well.\\
  User: Yeah, me too.\\
  SYS: What other Youtube compilations do you watch?\\
  User: Fails, and cats\\
  SYS: Cats are too cute, do you own one?\\
  User: No, maybe one day.\\
  SYS: Well I hope you will have a cat one day.\\
  User: Can you get me one?\\
  SYS: Unfortunately I can’t, but I can say purr purr.
  
\section{Architecture}
\textbf{Prompt:} Describe, in detail, your scientific approach along with the related system architecture. Plan for operating at scale when you define the architecture.
	\begin{figure}[!ht]
		\centering
		\includegraphics[width=0.95\textwidth]{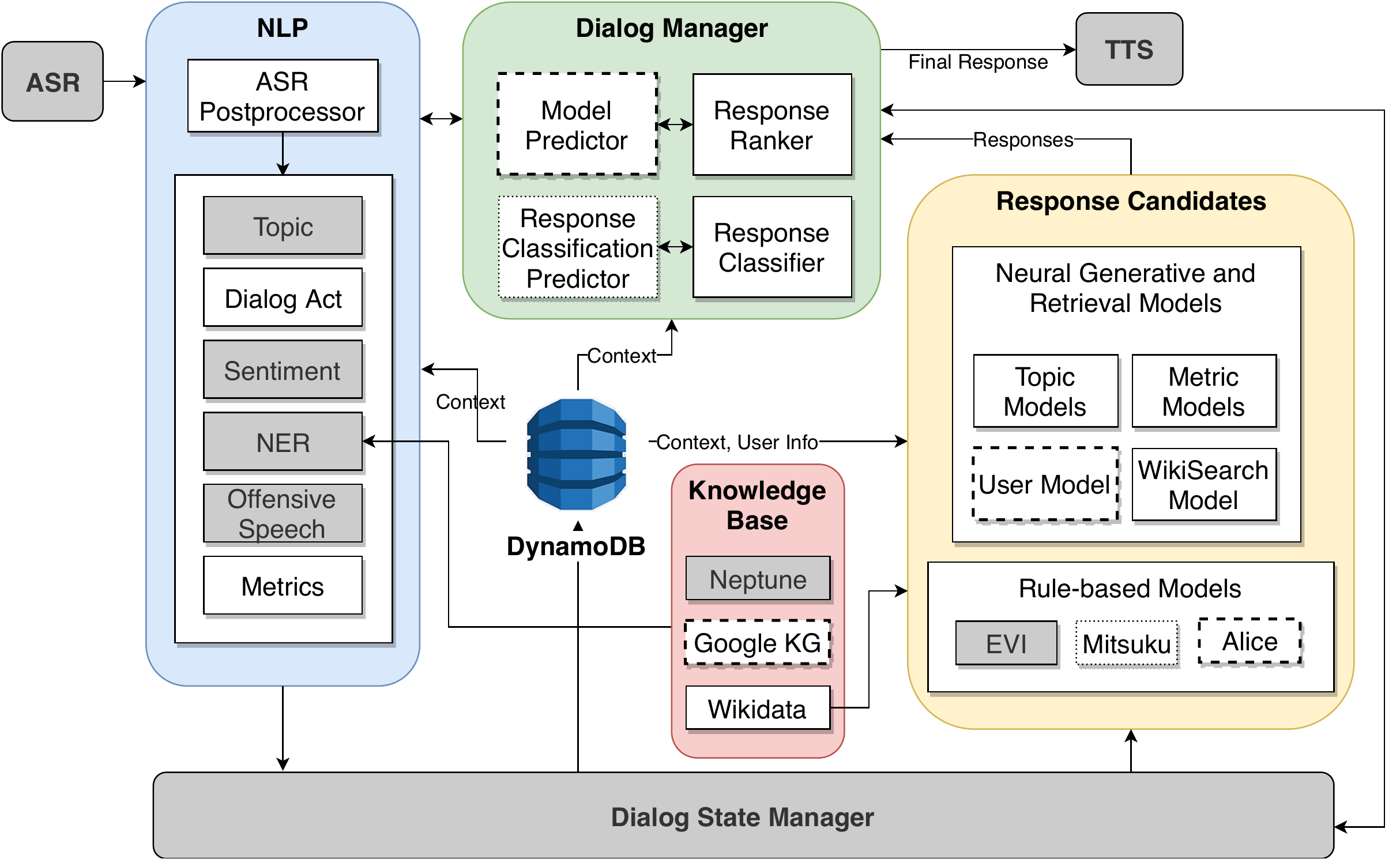}
		\caption{System architecture. Components with gray background are provided by Amazon. Components marked \texttt{core}, \texttt{core+}, and \texttt{core++} are to be completed by the end of phase 5, 7, and 9, denoted by a solid, long dotted, and short dotted outline, respectively.}
		\label{fig:arch}
	\end{figure}

Our system architecture (Figure~\ref{fig:arch}) is comprised of 3 main components: First the user input is processed through the \textcolor{myblue}{\bf NLP} component, then this data is sent to \textcolor{myyellow}{\bf Response Candidates}, which contains a suite of neural and rule-based models, and finally the \textcolor{mygreen}{\bf Dialog Manager} chooses the best response taking into account dialog history and current \textcolor{myblue}{\bf NLP} data. These components communicate with each other through the \textcolor{mygray}{\bf Dialog State Manager}, and a \textcolor{myred}{\bf Knowledge Base} component that can be queried by our components stores our knowledge bases.
We build on top of CoBot \cite{Khatri:2018}, thus the system is scalable and new components can be added easily. We leverage former Alexa competitors' architectures \cite{Chen:2018a, Serban:2017a}.
We minimize latency, by running tasks in parallel whenever possible, in order to make the conversation feel natural. Some redundancy is also included (e.g. in the form of multiple response generators), and we define a fixed time window for each major step in our pipeline, after which we interrupt the current component and use the information already computed from the sub-components in the next step, reducing total processing time.
We will develop our system in three phases (Figure~\ref{fig:arch}):
Components marked \texttt{core}, \texttt{core+}, and \texttt{core++} are to be completed by the end of phase 5, 7, and 9, respectively.
These are the minimally planned components for each category, but if time permits we will advance faster. This provides us an incremental and iterative approach to build our architecture starting with the most important components, always testing included components before advancing to new ones.
Our main novelties\footnote{Novelties are described in more detail in the document about novel approaches.} include:
\begin{itemize}[topsep=2pt,itemsep=-1ex,partopsep=1ex,parsep=1ex]
	\item Using self-play strategies to train a neural response ranker.
	\item Computing a large number of metrics for both input and response, and specifically optimizing some models for our metrics.
	\item Training a separate dialog model for each user.
	\item Using a response classification predictor and a response classifier to predict and control aspects of responses.
	\item Predicting which model emits the best response before response generation.
	\item Using our entropy-based filtering approach to filter dialog datasets \cite{Csaky:2019}.
	\item Using big, pre-trained, hierarchical BERT and GPT models \cite{Devlin:2018,Radford:2019,Wolf:2019}.\\

\end{itemize}
Next, we describe each component in detail in order of the data flow.
\paragraph{Dialog State Manager.} This is included in CoBot and we extend it to manage our current dialog state (i.e. conversations and related data described below), saving it to DynamoDB \cite{Decandia:2007} when appropriate. DynamoDB stores all past dialog states for every user. The \textcolor{gray}{\bf Dialog State Manager} communicates with the \textcolor{myblue}{\bf NLP} and \textcolor{mygreen}{\bf Dialog Manager} components which can update the dialog state. It works in parallel to all components, thus it doesn't affect latency.

\paragraph{NLP.} \textcolor{mygray}{\bf ASR} data is sent to the first component in the pipeline (\textcolor{myblue}{\bf NLP}), starting with the \textcolor{myblue}{\bf ASR Postprocessor}. If the confidence score of the transcribed utterance is below a certain threshold the pipeline is interrupted and we return a reply asking the user to repeat or rephrase their sentence. Otherwise if the confidence is above this but still lower than average we look at the n-best \textcolor{mygray}{\bf ASR} hypotheses and try to correct the utterance based on context (planned to be part of \texttt{core++}.). The corrected utterance is passed to all the subcomponents running in parallel. Token-timing is also saved to the dialog state and used as additional input to dialog models, as it might help disentangle separate phrases. We leverage and extend some of CoBot's built-in NLP components (\textcolor{myblue}{\bf NER}, \textcolor{myblue}{\bf Topic}, \textcolor{myblue}{\bf Sentiment}, and \textcolor{myblue}{\bf Offensive Speech} classifiers) and also add our own. Named entities are extracted and we use \textcolor{myred}{\bf Neptune} \cite{Bebee:2018} and the \textcolor{myred}{\bf Google Knowledge Graph} \cite{Chah:2018} to get related entities and pieces of information about them\footnote{This is planned to be part of \texttt{core+}.}. \textcolor{myblue}{\bf Topic}, \textcolor{myblue}{\bf Dialog Act}, \textcolor{myblue}{\bf Sentiment} and \textcolor{myblue}{\bf Offensive Speech} classifiers take into account previous dialog states (context) from DynamoDB. We save all information in DynamoDB and build statistics about the user (e.g. what are her/his favorite topics). We compute all our automatic evaluation metrics \cite{Csaky:2019} for the user utterance which is useful for the response selection strategy (e.g. if we find the user is bored we would try to suggest a new topic based on saved user information). After all subcomponents are finished or the time window is exceeded, all data is sent to the \textcolor{mygray}{\bf Dialog State Manager}. We also plan to experiment with inserting a response classification prediction (\textcolor{mygreen}{\bf RCP}) step\footnote{This is planned to be part of \texttt{core++}.}, which predicts the topic, dialog act and sentiment of the response, using context, and current \textcolor{myblue}{\bf NLP} data. The predicted information about the response is added to the dialog state and the dialog models in \textcolor{myyellow}{\bf Response Candidates} can leverage it. We also plan to experiment with using this information to control desired aspects of the response\textsuperscript{3} \cite{Xu:2018a}.
\paragraph{Response Candidates.} Once the \textcolor{myblue}{\bf NLP} and \textcolor{mygreen}{\bf RCP} are done, the \textcolor{mygray}{\bf Dialog State Manager} sends the current dialog state to our dialog models running in parallel. Most models will also use conversation history and user information from DynamoDB.
Ensemble modeling, a prevalent technique in nearly all Alexa socialbots \cite{Serban:2017a, Khatri:2018}, improves the response quality since we can have different models dealing with different domains and situations. Rule-based models include \textcolor{myyellow}{\bf Evi} (built into CoBot), and publicly available AIML parts of \textcolor{myyellow}{\bf Alice}\footnote{This is planned to be part of \texttt{core+}.} \cite{Abushawar:2015} and \textcolor{myyellow}{\bf Mitsuku}\footnote{\url{https://www.pandorabots.com/mitsuku/}}\textsuperscript{,3}. The base of all neural models is a big, hierarchical BERT or GPT-based model, pre-trained on non-dialog data \cite{Wolf:2019,Serban:2015}. The hierarchical part ensures that our models are grounded in past utterances and that they respond differently to the same input utterance (since the past is different). We also plan to experiment with inserting BERT layers in variational models\textsuperscript{3} \cite{Park:2018,Gu:2019}, which can provide more interesting and non-deterministic responses. We further train our pre-trained models on all available dialog datasets jointly. Finally, we finetune \textcolor{myyellow}{\bf Topic Models} on datasets related to specific topics (e.g. subreddits), while \textcolor{myyellow}{\bf Metric Models} are finetuned jointly on all dialog datasets, but we replace the loss function with a specific metric (e.g. coherence, diversity, etc.). \textcolor{myyellow}{\bf Metric Models} can focus on specific dialog properties and ensure that generated responses are diverse and engaging. We train models with extra annotations (e.g. topic, sentiment in DailyDialog \cite{Li:2017b}, or using knowledge pieces \cite{Dinan:2018} through the new Amazon topical chat dataset). There are several issues with the cross-entropy loss function \cite{Csaky:2017,Csaky:2019,Li:2016d,Wei:2017,Shao:2017b,Zhang:2018a,Wu:2018}, and we proposed to use all kinds of features \cite{Csaky:2017}, motivating the usage of annotations computed with \textcolor{myblue}{\bf NLP}, which helps amend the loss function problem and provides more interesting and diverse responses \cite{Zhou:2018,Li:2017b,Xing:2017,Liu:2017,Baheti:2018,Dinan:2018,Ghazvininejad:2018,Zhu:2017}. We use two variants of each \textcolor{myyellow}{\bf Topic} and \textcolor{myyellow}{\bf Metric Model}, a neural generative and a retrieval based, which simply returns the n-best responses from training data. The \textcolor{myyellow}{\bf User Model}\textsuperscript{4} is a user-specific dialog model finetuned on user-Alexa conversations. It will be at least one order of magnitude smaller than other models since we have to train and store the weights (in DynamoDB) of one model for each user. Through this model, we can encode information about the user, and the model can stay more consistent (if trained with its own responses as targets). Personalizing our system is important and we feel that it will make our chatbot more pleasant to talk to \cite{Zhou:2018a}.
The \textcolor{myyellow}{\bf WikiSearch Model} simply searches \textcolor{myred}{\bf Wikidata} \cite{Vrandevcic:2014} and returns relevant sentences which we can consider as responses. A similar model is employed for the Washington Post live API as well to stay up-to-date with events and news.
We also plan to experiment with an ensemble model setup, where all the response candidates are combined into one response word-by-word, which can be considered as an additional response candidate.
Through the \textcolor{myyellow}{\bf User Model} and the knowledge-augmented \textcolor{myyellow}{\bf Topic Models} our goal is to achieve an engaging and interesting conversation in which topic handling and topic switching occur more naturally than in current Alexa socialbots. In the initial stages of the competition, we plan to experiment with as many models as possible and use crowdsourcing to exclude from our system models that generate low-quality responses.
\paragraph{Dialog Manager.} Once all dialog models have computed a response or timed out, we send response candidates to the \textcolor{mygreen}{\bf Dialog Manager}. The \textcolor{mygreen}{\bf Model Predictor}\textsuperscript{4} runs in parallel with the dialog models, trying to predict which model will generate the best response based on the dialog state and context. If we find that such a model can predict the selected model (by the \textcolor{mygreen}{\bf Response Ranker}) accurately, then we can largely decrease computational costs by reducing the number of models required to produce a response. We plan to experiment with several response selection strategies (\textcolor{mygreen}{\bf Response Ranker}) and evaluate them with crowdsourced evaluators in A/B tests. In the initial phases (\texttt{core} part) of the competition, we plan to employ safe baseline strategies like selecting responses only from retrieval and rule-based models, using the CoBot selection strategy, and ranking responses using a weighted sum across all metrics. Our end goal is to be able to learn a neural ranker, which takes as input the dialog state, context, and response candidates (and their probability scores in the case of neural models), and outputs the best response\textsuperscript{4}. One approach is to use crowdsourcing to gather training data for the ranker, by letting people choose the best response among the candidates. We also plan to use user feedback with reinforcement learning \cite{Serban:2017a}. In the final version of the ranker we plan to experiment with self-play\textsuperscript{3} \cite{Silver:2017,Burda:2018,Xu:2018a}, described in detail in the novel approaches document of the application. Essentially, both at train and test times we can do rollouts with the ranker, where the dialog system feeds its response into itself, to filter responses that lead to poor conversations. This is a computationally taxing technique, which will be tuned to the desired latency. To increase selection confidence we will use the agreement between the \textcolor{mygreen}{\bf Model Prediction} and selected response, and between the \textcolor{mygreen}{\bf Response Classifier} and the \textcolor{mygreen}{\bf RCP}. The \textcolor{mygreen}{\bf Response Classifier} uses the \textcolor{myblue}{\bf NLP} component to compute the same data as the \textcolor{mygreen}{\bf RCP}, and is useful in helping the \textcolor{mygreen}{\bf Response Ranker} rank responses, based on whether they are offensive, on topic, positive, engaging, etc., ensuring a fun and interesting conversation. Thus the \textcolor{mygreen}{\bf Response Ranker} leverages all components in the \textcolor{mygreen}{\bf Dialog Manager} before emitting the final response, which is sent to the \textcolor{mygray}{\bf TTS} of Alexa. The \textcolor{mygreen}{\bf RCP} and the \textcolor{mygreen}{\bf Model Predictor} are both trained so they approximate their post-\textcolor{myyellow}{\bf Response Candidates} counterparts (\textcolor{mygreen}{\bf Response Classifier} and \textcolor{mygreen}{\bf Response Ranker}). This training signal can be used in the loss function of the neural dialog models as well.

\newpage
\section{Novelty}
\textbf{Prompt:} What is novel about the team’s approach? (This may be completely new approach or novel combination of known techniques).\\
\\
Our novelties include:
\begin{enumerate}
\item Using self-play learning for the neural response ranker (described in detail below).
\item Optimizing neural models for specific metrics (e.g. diversity, coherence) in our ensemble setup.
\item Training a separate dialog model for each user, personalizing our socialbot and making it more consistent.
\item Using a response classification predictor and a response classifier to predict and control aspects of responses such as sentiment, topic, offensiveness, diversity etc.
\item Using a model predictor to predict the best responding model, before the response candidates are generated, reducing computational expenses.
\item Using our entropy-based filtering technique to filter all dialog datasets, obtaining higher quality training data \cite{Csaky:2019}.
\item Building big, pre-trained, hierarchical BERT and GPT dialog models \cite{Devlin:2018,Radford:2019,Wolf:2019}.
\item Constantly monitoring the user input through our automatic metrics, ensuring that the user stays engaged.
\end{enumerate}

Self-play \cite{Silver:2017,Burda:2018} offers a solution to the scarcity of dialog datasets and to the issues encountered when using cross-entropy loss as an objective function \cite{Csaky:2017,Csaky:2019}. In our setup the dialog system would converse with itself, selecting the best response with the neural ranker in each turn. After a few turns, we reward the ranker based on the generated conversation. Our reward ideas include a weighted sum of metrics and using crowdsourcing and user ratings. Furthermore, we wish to explore two exciting self-play setups: 1. An adversarial setup where the ranker is trained to generate a dialog by self-play to fool a neural discriminator deciding whether it’s machine or human generated. 2. We apply the ideas of curiosity and random network distillation to train the neural ranker \cite{Burda:2018}. We also plan to experiment with self-play ideas for some of the individual neural dialog models.

\newpage
\section{Related Work}
\textbf{Prompt:} Please provide a summary of the technical work/research (relevant to your proposed architecture), yours or others’, that you will leverage and how.\\
\\
We employ topic, dialog act, and sentiment classifiers, which widely used in the literature \cite{Chen:2018a}. We leverage rule-based bots in our system because they can provide a different class of responses than neural models. We use recent NLP models \cite{Devlin:2018,Radford:2019}, by finetuning them on our dialog datasets, and modify them to be more suited to deal with dialog modeling, e.g. making them hierarchical or integrating them in other state-of-the-art dialog models \cite{Park:2018,Gu:2019}. We leverage baseline response rankers, and adapt ideas from the domains of reinforcement and self-play learning to dialog modeling \cite{Serban:2017a, Xu:2018a,Burda:2018}.\\
\\
We and others have found the cross-entropy loss function problematic and the primary reason for the generation of short and boring responses \cite{Csaky:2017,Li:2016d,Wei:2017,Shao:2017b,Zhang:2018b,Wu:2018}. To amend this, we use our idea of filtering dialog datasets based on entropy, obtaining higher quality training data \cite{Csaky:2019}. We address the loss function problem using various features and metrics (from the NLP component) and knowledge pieces (using the new topical chat dataset and Wikidata), which can help neural models in generating more natural and diverse responses \cite{Zhou:2018,Li:2017b,Liu:2017,Baheti:2018,Dinan:2018,Zhu:2017}.\\
\\
We build on top of, modify, and extend CoBot and former competitors’ architectures, as they provide a solid foundation for our dialog system \cite{Chen:2018a,Serban:2017a}. ASR postprocessing, and a neural ranker choosing between response candidates are some techniques that we include in our architecture.

\newpage
\section{Ensuring an engaging experience}
\textbf{Prompt:} How will you ensure you create an experience users find engaging?\\
\\
We have several mechanisms to ensure an engaging experience:
\begin{enumerate}
\item We classify user utterances by topic, sentiment, etc., and calculate our automatic metrics, using this information when selecting and generating responses. If we find the user lost interest in the conversation, we might suggest a new topic related to their interests (through our topic and user models).
\item We also classify the response candidates, so that we can make sure that they are engaging and relevant. With the help of knowledge-augmented models \cite{Dinan:2018} we offer the user an interesting and informative conversational experience in a natural way, which all contributes to engagingness. 
\item Personalization through our user models is an important factor to engagingness. If the user feels like the socialbot is able to remember and include past information about them in its responses, then this directly contributes to building a deeper connection with the user and maintain their interest. 
\item Defining a maximum latency for our pipeline is a small but important feature to ensure users stay engaged.
\item We plan to heavily leverage user feedback (by classifying verbal feedback and using ratings to refine our response ranker) in order to improve our system.
\end{enumerate}

\newpage
\section{Impact}
\textbf{Prompt:} How do you think your work will impact the field of Conversational AI?\\
\\
We aim to move forward the field of conversational AI and neural dialog modeling through three main novelties: self-play learning, tackling the loss function problem, and personalization.\\
\\
We believe that instead of using rule-based components and rule-based dialog managers, our refined, self-play based, neural ensemble system is much more capable of scaling and will be a great step forward in the field towards achieving a better conversational experience. Our work will popularize the idea of self-play and will eliminate some of the problems with current neural dialog models. We believe that applying self-play on the response ranker level is an under-researched idea, with which we could potentially train much better dialog agents than current ones.\\
\\
We believe that our combined approaches will mend the problems with learning through the cross-entropy loss function, and will create more diverse and interesting dialog models \cite{Csaky:2017,Csaky:2019}. These include feeding classification and metric annotations to our neural models, optimizing models specifically for our metrics, and using self-play.\\
\\
From a user perspective, personalization in dialog agents is one of the most important aspects that current socialbots are lacking, which we want to make a great impact on through our user models.

\newpage
\bibliography{../../../Common/Bib/ml}
\bibliographystyle{apalike}
	
\end{document}